\documentclass[11pt]{article}

\usepackage[final]{acl}

\usepackage{times}
\usepackage{latexsym}

\usepackage[T1]{fontenc}

\usepackage[utf8]{inputenc}

\usepackage{microtype}

\usepackage{inconsolata}
\usepackage{hyperref}
\usepackage{graphicx}

\usepackage{amsmath}
\usepackage{amssymb}
\usepackage{xcolor}
\usepackage{booktabs}
\usepackage{multirow}
\definecolor{darkgreen}{RGB}{0, 150, 0}
\newcommand{\newtext}[1]{\textcolor{black}{#1}}

\title{Can Active Label Correction Improve LLM-based Modular AI Systems?}

\author{Karan Taneja \\
  Georgia Institute of Technology \\
  \texttt{karan.taneja@cc.gatech.edu} \\\And
  Ashok K. Goel \\
  Georgia Institute of Technology \\
  \texttt{ashok.goel@cc.gatech.edu} \\}

\begin{document}
\maketitle
\begin{abstract}

Modular AI systems can be developed using LLM-prompts-based modules to minimize deployment time even for complex tasks.
However, these systems do not always perform well and improving them using the data traces collected from a deployment remains an open challenge.
The data traces contain LLM inputs and outputs, but the annotations from LLMs are noisy. 
We hypothesize that Active Label Correction (ALC) can be use on the collected data to train smaller task-specific improved models that can replace LLM-based modules. 
In this paper, we study the noise in three GPT-3.5-annotated datasets and their denoising with human feedback.
We also propose a novel method \textbf{ALC3} that iteratively applies three updates to the training dataset: auto-correction, correction using human feedback and filtering.
Our results show that ALC3 can lead to oracle performance with feedback on 17-24\% fewer examples than the number of noisy examples in the dataset across three different NLP tasks.

\end{abstract}

\section{Introduction}

Large language models (LLMs) such as OpenAI's ChatGPT \cite{openai_introducing_2022}, Google's Bard based on LaMDA \cite{thoppilan_lamda_2022}, and Meta's LLaMA \cite{touvron_llama_2023} are powerful zero or few-shot learners as they can generalize to a wide range of tasks simply through task explanations or demonstrations without any model fine-tuning \cite{qin_is_2023}. 
This led to the development of many modular AI systems that leverage these LLMs through API services to decompose complex tasks and to automate workflows \cite{shen_hugginggpt_2023, significant_gravitas_auto-gpt_2023, tellez_these_2023,taneja_jill_2024}. 
These AI systems rely on several discriminative and generative sub-tasks performed by LLMs, without any fine-tuning, to achieve the desired behavior. 
LLM-based modular AI systems work reasonably well in practice, but their robustness is still in question because (i) they are limited by the quality of zero-shot learning of the LLMs that they employ \cite{shen_hugginggpt_2023}, 
(ii) an error in one module can trigger a cascade effect, propagating the error throughout the modular system,  
(iii) AI developers cannot rely on future improvements in the quality of underlying LLMs as the performance on certain end-tasks can also worsen \cite{chen_how_2023} due to the fine-tuning focused on safety and conversationality of LLMs, 
and
(iv) domain shift between LLM training data and deployed environment.  
Further, LLMs cannot be easily fine-tuned for all sub-tasks in every AI system because 
(i) there is a lack of training data, and 
(ii) the computational cost required to train and maintain an online service for multiple fine-tuned models can be expensive and has a large carbon footprint. 

\begin{figure}[t]
    \centering
    \includegraphics[width=0.99\linewidth]{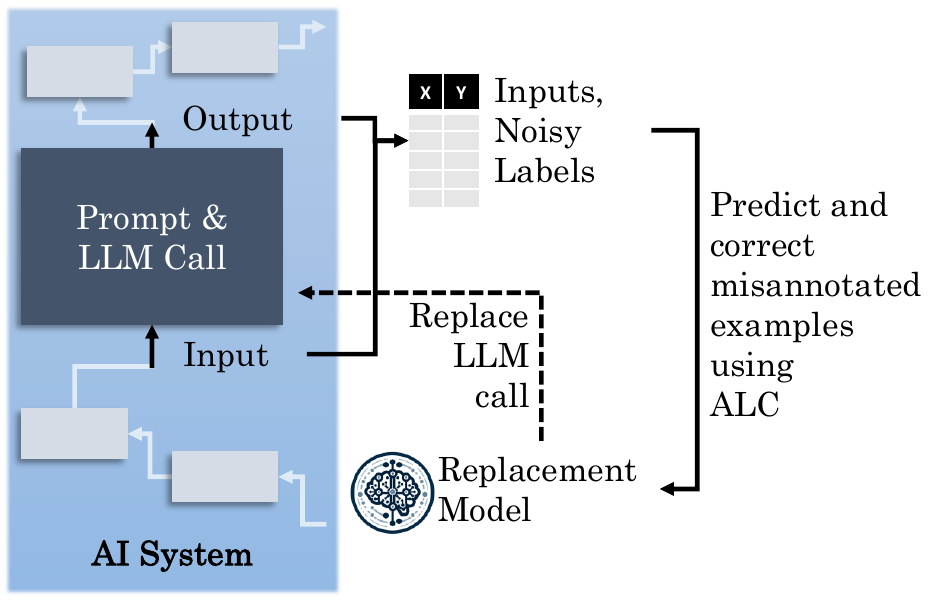}
    \caption{Noisy LLM-annotated datasets are collected from deployment of a modular AI system. Active Label Correction (ALC) is used to predict and correct misannotated examples in order to train a replacement model.}
    \label{fig:teaser}
\end{figure}

To address these challenges, we propose collecting data traces containing LLM inputs and outputs from deployed AI systems, and training smaller task-specific models that can replace LLM calls (see Figure \ref{fig:teaser}).
We propose a method based on Active Label Correction (ALC) \cite{rebbapragada_active_2012} to use human feedback to improve the quality of data obtained from modular AI systems  and evaluate its efficacy for three publicly-available NLP datasets.
The data traces contain examples with low-quality LLM annotations, but ALC can utilize human expertise in an efficient manner by seeking annotations for examples that are most likely misannotated.
We show that synthetic noise models employed by previous works are not adequate to capture noisy LLM annotations which makes it challenging to predict misannotated examples in ALC.
The proposed method \textbf{ALC3} (i) auto-corrects potentially misannotated examples using model predictions, (ii) filters out the confusing examples in addition to (iii) using human feedback.
\textit{Perfect} misannotation prediction on a dataset with $N\%$ noisy examples will require human annotation on $N\%$ of data to clean it. 
Our results show that the three-step process of \textbf{ALC3} can lead to oracle test performance with feedback on 17-24\% fewer examples than $N\%$ (perfect predictor) for three NLP tasks with different data sizes (5k-100k) and complexity.
The proposed method promotes the idea of leveraging the powerful zero-shot learning capabilities of LLMs for rapidly deploying modular AI systems and improving their quality as they accumulate data from the real world.
We focus on discriminative tasks but the proposed method also extends to generative tasks in principle (see \S\ref{appendix:extension-generative}).

We make three main contributions: 
(i) we propose an ALC-based pipeline \newtext{and our method \textbf{ALC3}} to continually improve LLM-based modular AI systems while minimizing annotation costs, 
(ii) we compare different noise models with LLM-annotation noise and the performance-cost curves of different ALC methods for GPT-3.5-annotated datasets,
and 
(iii) \newtext{we present experiments for three publicly-available NLP datasets with tasks} that are common to modular AI systems including natural language inference, named entity recognition, and intent classification.

\section{Related Work}

\textbf{LLM-based Modular AI Systems}:
ChatGPT \cite{openai_introducing_2022}, LaMDA \cite{thoppilan_lamda_2022} are conversational LLMs trained on large amounts of text and fine-tuned with human feedback, and they perform well on diverse NLP tasks but not at par with fine-tuned models \cite{qin_is_2023}.
HuggingGPT \cite{shen_hugginggpt_2023}, AutoGPT \cite{significant_gravitas_auto-gpt_2023}, Jill Watson \cite{taneja_jill_2024} are examples of modular AI systems based on GPT-3.5 to solve complex tasks like planning, goal setting, manipulating images or texts.
These AI systems depend on LLMs as API services and are expensive to run.
We propose a pipeline to use data traces from the deployment of AI systems and limited human supervision to improve their performance and to make them cost-effective by using smaller models.

\begin{figure*}[]
    \centering
    \includegraphics[width=\linewidth]{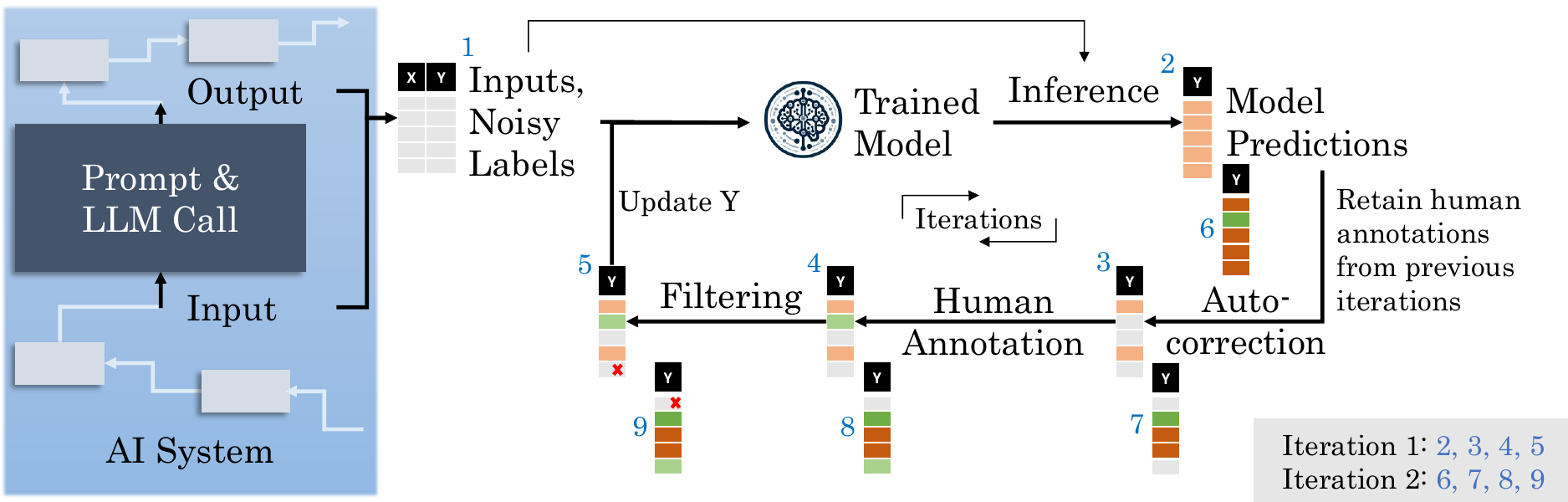}
    \caption{
    \textbf{Proposed process for improving LLM-based modular AI systems using ALC3.} 
    The \textit{inputs} and \textit{noisy labels} from a zero/few-shot learner-based module are used to obtain a \textit{trained model}. 
    \textit{Model predictions} on the noisy training dataset are computed for the next three steps. 
    (i) \textit{Auto-correction} updates the labels where model prediction contradicts the original label with very high confidence. 
    (ii) \textit{Human annotation} is used to verify and update a fixed number of confusing examples.
    (iii) \textit{Filtering} removes some of remaining examples that are deemed noisy based on model predictions.
    The process is performed iteratively until a stopping condition.
    Only human annotations are retained after each iteration, iteration two is shown with columns 6, 7, 8, and 9 for illustration.
    }
    \label{fig:main}
\end{figure*}

\textbf{Semi-supervised, Weakly-supervised and Active Learning:}
In semi-supervised learning, training data consists of a set of labeled examples and a typically much larger set of unlabeled examples \cite{hady_semi-supervised_2013}.
Weakly supervised learning more generally assumes that annotations are partial, noisy, or provided at a higher level of abstraction \cite{zhou_brief_2018}.
Self-training is a common method used to train a teacher model on the supervised subset to predict soft pseudo-labels for the unsupervised subset. 
Then, the supervised subset and pseudo-labeled subset are used to train a student model \cite{karamanolakis_self-training_2021,tanaka_joint_2018,yi_probabilistic_2019}.
Active learning strategy selects the most informative examples, typically by construction or by choosing from a pool of unlabeled examples, and querying an oracle such as human annotator for supervision \cite{settles_active_2009}. 
Our problem is different from above but lies at the intersection since we begin with a fully-annotated noisy dataset and need to seek human expertise for improving data and model quality.

\textbf{Active Label Correction (ALC):}  ALC \cite{rebbapragada_active_2012} uses self-training to clean a noisy training set by iteratively predicting most likely mislabeled examples using a trained model and receiving human feedback to correct them. 
Simple disagreement measures based on the output probabilities work well for identifying mislabeled examples. 
Robust ALC \cite{kremer_robust_2018} assumes that label-conditioned noise and uses a latent model for the true label.
Dual ALC or DALC \cite{li_improving_2022} additionally uses model predictions with high confidence for automatic correction.
We note that previous works have primarily focused on image classification problems in vision and experimented with synthetic noise.
We study and explore the use of ALC for correcting real noise observed in GPT-3.5-annotated datasets.

\textbf{Human-AI Collaborative Annotation:} CoAnnotating \cite{li_coannotating_2023} is a data annotation process that uses variations in LLM prompts to estimate model uncertainty for deciding allocation to annotators.
In our work, we boost data quality towards the goal of continual model improvement to replace LLM-based modules in AI systems.
As a result, we use an iterative process where human feedback helps increase model and misannotation prediction quality i.e. better prediction of examples most likely to be misannotated, along with auto-correction and filtering. 
CoAnnotating is non-iterative and does \textit{not} leverage fine-tuned models or previously annotated human examples, 
but relies on numerous LLMs calls for each example.
Machine Teaching \cite{taneja_framework_2022, taneja_human-ai_2022} is another framework for human-AI collaborative annotation where annotators work with an ML model and knowledge base to \textit{teach} concepts to machines in a selective labeling process. 
In our proposed process, LLMs can be interpreted as a source of knowledge as they provide initial annotations, but the presence of annotations transforms the problem into one of label correction.  

\section{Methods}
Modular AI systems 
rely on the zero/few-shot learning power of LLMs to process the input, perform sub-tasks, and generate output.
These AI systems can accumulate data traces that include inputs, intermediate outputs of each module, and final outputs through user interactions.   
To evaluate a module that relies on LLMs as a zero-shot learner, one can analyze the amassed inputs and outputs. 
We propose collecting these input-output pairs for each module to create noisy annotated datasets, and improving data quality using human feedback with the goal of training a replacement model. 
The datasets above are assumed to be noisy because LLM annotations have lower quality than human annotations. 

ALC \cite{rebbapragada_active_2012} iteratively predicts and flag the examples that are most likely to be noisy or misannotated, and seeks correction from human annotators on the flagged examples. 
The iterative approach aims to jointly
(i) minimize noise in training examples which is also a proxy for test error, and 
(ii) minimize human feedback by effectively predicting misannotated examples. 
Our proposed method \textbf{ALC3}, outlined in Figure \ref{fig:main}, uses two additional steps beyond the basic ALC.
We first auto-correct the examples where model prediction contradicts the label with very high confidence, leveraging the trained model for labelling, similar to Dual ALC \cite{li_improving_2022}.
After human annotation, we filter out examples with low label probabilities to reduce the noise in the training dataset.
In each iteration, we retain the human annotations, but reset and obtain a new set of auto-corrected and filtered examples.
We describe each step in more details in the following subsections.

\subsection{Misannotation Prediction}

The task of \textit{misannotation prediction} (MP) involves predicting the examples that are most likely to be noisy based on the entire noisy dataset.
This ensures efficient \textit{human annotation} as annotators focus on examples most likely to improve the dataset and the trained model. 
Given a noisy dataset $\mathcal{D}=\{(x_i,y_i)\}_{i=1}^{n}$ where $(x_i, y_i)$ are pairs of inputs and outputs, we train a model $p_\theta$ to predict $y^*=\arg \max_y p_\theta (y|x)$. 
We then obtain predictions on the $\mathcal{D}$ itself and compute probabilities $p_\theta(y_i|x_i)$ for each example. 
Similar to \citet{rebbapragada_active_2012},
we suppose that an example is misannotated with probability $m(x_i, y_i) = 1 - p_\theta(y_i|x_i)$. 
In other words, if the probability of $(x_i, y_i)$ is low under the trained model, it is likely to be misannotated ($m \approx 1$) and vice versa.

Intuitively, when model prediction for a training example is inconsistent with the annotation, we can argue that the remaining examples suggest evidence against the given annotation and, therefore, it is likely to be misannotated.
Since each example has a minimal impact on the model for a large enough sample size, $p_\theta(y_i|x_i)$ approximates as the probability of the model trained with one example removed from the dataset. 
We study this impact of data size in the next section.

After computing $m(x_i, y_i) $ for all  $(x_i, y_i) \in \mathcal{D}$, we sort $(x_i, y_i) \in \mathcal{D}$ from highest to lowest value of $m(x_i, y_i)$. 
We flag a fraction $M \in (0, 1)$ of examples in $\mathcal{D}$ with the highest $m(.)$ values for human annotation.
A low value of fraction $M$ leads to a low recall as many misannotated examples are missed while a high $M$ leads to lower precision because many correctly annotated examples are flagged.
We refer to this baseline as \textbf{ALC}.
We also evaluate Random Label Correction (\textbf{RLC}) as a baseline where the examples are flagged randomly. 

\subsection{Auto-correction}
While ALC only employs human supervision for correction, the model predictions can also be a source of correct annotations. 
Dual ALC \cite{li_improving_2022} leverages model predictions along with human feedback to update the dataset.
Instead of using label-transition matrix for noise modeling, we adapt their method by simply using model predictions to update labels before determining misannotated examples for direct comparison with other ALC methods. 
We accept output label $y^*$ as true annotation for an example if $p_{\theta}(y^*|x) > \delta$ where $\delta \lesssim 1$ is a constant threshold.
We refer to this as \textit{auto-correction} since it requires no human input.
We use remaining examples to correct the most likely misannotated examples as before
and refer to this baseline method as \textbf{DALC}.
Since auto-correction is performed before human annotation, we do not flag any examples that are auto-corrected.

\subsection{Filtering}

\newcommand{\mflagged}{m_{\text{flag}}}
\newcommand{\mcorrected}{m_{\text{corr}}}
\newcommand{\MPPrecision}{p_{\text{MP}}}
\newcommand{\mfiltered}{m_{\text{filter}}}

In \textbf{ALC3}, after auto-correction and human annotation, we filter out the examples that are likely to be noisy to improve the quality of the training data.
Let $\mflagged$ be the number of  flagged examples using MP and $\mcorrected$ be the number of corrected examples in the current iteration.
Then, MP precision is given by $\MPPrecision = \mcorrected / \mflagged$.
In our experiments, we filter out $\mfiltered$ examples with lowest values of $p_\theta(y|x)$ where $\mfiltered = 3  \mcorrected$, a rough approximation of number of examples that will be corrected in the next few iterations.

As $\MPPrecision$ drops with each iteration, effectiveness of filtering in terms of lowering the noise in training dataset reduces. 
We estimate the noise in every iteration and perform filtering only if $\MPPrecision$ is strictly better than random selection. Therefore, for $k^{th}$ iteration, we use
$$ 
\mfiltered{_{,k}} = 
\begin{cases} 
3 \cdot \mcorrected{_{,k}} & \text{if } \MPPrecision{_{,k}} > \eta_{k} \\
0 & \text{otherwise} 
\end{cases} 
$$
$$\eta_{k} = \eta_o - \frac{ \sum_{j = 1}^{k} \mcorrected{_{,j}} }{ |\mathcal{D}| }$$
where $\eta_o$ is the fraction of noisy examples initially which is estimated by annotating a small random subset that is also used as the test set.  

\subsection{Iterative Data Correction}

For each iteration of \textbf{ALC3}, we first train $p_\theta$, and perform auto-correction using $p_\theta(y^*|x)$ as discussed earlier. This is followed by misannotation prediction using $p_\theta(y|x)$ where examples are flagged for human annotation.
Once the flagged examples are annotated by a human expert, some remaining examples are filtered out to further improve data quality based on low $p_\theta(y|x)$.
Finally, after above three updates, this dataset is used to train a new model $p_{\theta'}$ for the next iteration. 
In each iteration, a new set of examples are auto-corrected, human-annotated, and filtered. 
We reset the previously auto-corrected or filtered examples and select a new set based on $p_{\theta'}$ predictions.
An human-annotated example is \textit{not} auto-corrected or flagged or filtered in subsequent iterations.
The process is repeated iteratively improve the quality of the data and trained model until a stopping criterion (discussed in Section \ref{subsection:results-iterative}) is met. 

With each iteration, new feedback from the human annotator is accommodated into the dataset and trained model. 
The new model predictions reveal new inconsistencies with the noisy annotations and lead to the next set of flagged examples. 
As the count of misannotated examples reduces in the noisy dataset with each iteration, we can expect the MP performance to drop.  
We discuss this phenomenon in the results section.
We can control the MP performance by varying $M$, the fraction of flagged examples.
Ideally, one can train a new model after correcting every example but this extreme case is expensive and infeasible due to long waiting time as model training runs after every example. 
But we still wish to choose $M$ to maximize MP precision and human efficiency. 
Therefore, in practice, the MP precision can be measured for the latest batch of examples and $M$ can be decreased if precision is low or vice versa.
Increasing precision by lowering $M$ comes at the cost of low recall and running more iterations.
In our experiments, we use a fixed value of $M$ i.e. a fixed number of examples are flagged in each iteration.

\section{Experiments and Results}

\subsection{Performance on Evaluation Tasks}

We evaluate the ALC-based pipeline on three different tasks: (1) \textbf{ATIS} (Airline Travel Information System) Intent Classification \cite{hemphill_atis_1990}, (2) \textbf{CoNLL} 2003 Named Entity Recognition or NER \cite{tjong_introduction_2003} and (3) Question-answering Natural Language Inference or \textbf{QNLI} \cite{wang_glue_2018}. 
The three tasks are common to many conversational AI agents \cite{lin_review_2023}.
Intent classification is used for processing inputs down to different path,
NER is used for slot-filling (such as train booking) to fill various parts of a database query, and
NLI is used in conversational AI agents to confirm the validity of the output in answering the user query.
The three datasets span different levels of complexity and sizes as shown in Table \ref{tab:ChatGPTPerf}.
The ground truth labels allow us to compare our pipeline to the oracle performance. 

For each task, we created a prompt on OpenAI Playground and used the OpenAI API to get outputs for both train and test splits of each dataset using `gpt-3.5-turbo-0613' model (Sept'21 cut-off) with zero temperature.
We use GPT-3.5 as it is available as an API service and performs well at following instructions. 
For fine-tuning, we used `distilbert-base-uncased', `roberta-base' and `albert-large-v2' pre-trained transformer models from HuggingFace for ATIS, CoNLL and QNLI tasks respectively based on their public leaderboards\footnote{\href{https://paperswithcode.com/}{https://paperswithcode.com/}} and availability on HuggingFace.
The results for each task are shown in \textbf{FT} (fine-tuned) column in Table \ref{tab:ChatGPTPerf}. 
The training details are given in \S\ref{appendix:training-details}.

\begin{table}[t]
    \centering
    \begin{tabular}{llccc}
    \toprule
    \multicolumn{1}{l}{\textbf{Task}} & \textbf{Metrics} & \textbf{Train} & \textbf{Eval} & \textbf{FT} \\
    \midrule
    \multirow{3}{*}{ATIS} & Size & 4,952 & 878 & 878 \\
     & Acc. & 0.702 & 0.756 & 0.982 \\
     & F1-sco. & 0.789 & 0.832 & 0.982 \\
    \midrule
    
    \multirow{5}{*}{CoNLL} & Size & 14,041 & 3,250 & 3,250\\
     &  \#NERs & 23,499 & 5,942 & 5,942 \\
     & Precision & 0.576 & 0.614 & 0.958 \\
     & Recall & 0.757 & 0.765 & 0.961 \\
     & F1-sco. & 0.654 & 0.682 & 0.959 \\
    \midrule
    
    \multirow{2}{*}{QNLI} & Size & 104,743 & 5,463 & 5,463\\
     & Acc. & 0.849 & 0.845 & 0.920 \\
    \bottomrule
  \end{tabular}
    \caption{Performance of GPT-3.5 on train and test set of three tasks: (1) {ATIS} Intent Classification, (2) {CoNLL} 2003 Name Entity Recognition, and (3) {QNLI} Natural Language Inference along with performance of fine-tuned models in \textbf{FT} column.}
    \label{tab:ChatGPTPerf}
\end{table}

ATIS dataset consists of 17 unique intents (`airport', `aircraft', `distance', etc.) assigned to user queries. 
A description of all 17 intents was provided to GPT-3.5 for classification. 
40 examples with multiple labels were removed to reduce the task to single-label classification.
For the CoNLL dataset, the prompt describes a sample NER output and four types of entities i.e. person, location, organization, and miscellaneous.
QNLI dataset consists of pairs of questions and sentences based on the Stanford Question Answering Dataset \cite{wang_glue_2018}. 
The prompt contains these two inputs and asks if the passage answers the question.
All prompts and output parsing are detailed in \S\ref{appendix:prompts}. 

The performance of GPT-3.5 is shown in \textbf{Train} and \textbf{Eval} columns in Table \ref{tab:ChatGPTPerf}. 
The results agree with \citet{qin_is_2023} in terms of the difference in performance between a fine-tuned model and zero-shot learning using GPT-3.
We note that GPT-3.5 has 10-30\% lower performance than fine-tuned models across the three tasks.
While GPT-3.5 generalizes well to new tasks as a zero/few-shot learner, this result suggests that there is a large performance gap that needs to be bridged when bootstrapping modular AI systems with GPT-3.5.
Therefore, it is important to develop a method to improve each module and the overall system over time after deploying the LLM-based AI systems. 

\subsection{Noise Characteristics}

We compare different types of synthetic noise models with the noise induced by GPT-3.5 while annotating the ATIS dataset. 
We use ATIS dataset for this experiment as it has the highest number of classes which can lead to more pronounced noise patterns.
We add different types of synthetic noise to the original dataset with the same proportion of misannotations as GPT-3.5. 
Table \ref{tab:noise-types} shows the accuracy of models trained with different noises in ATIS dataset and the class imbalance measured by the KL divergence of label distribution from the original label distribution. 
Details about noise models are provided in Appendix \ref{appendix:noise-models}.

\begin{table}[t]
    \centering
    \begin{tabular}{lcc}
        \toprule
        \textbf{Noise} & \textbf{Accuracy} & \textbf{C. Imbalance}\\
        \midrule
        None & 98.2\% & - \\
        Random & 97.2\% & \textbf{0.1697} \\
        Label-cond. & 93.5\% & 0.0485 \\
        Input-cond. & 88.6\% & 0.1033 \\
        GPT-3.5 & \textbf{84.0\%} & 0.0949 \\
        \bottomrule
    \end{tabular}
    \caption{Comparison of different noise types, including no noise (\textbf{None}), \textbf{Random}, \textbf{Label-cond}itional, \textbf{Input-cond}itional, and \textbf{GPT-3.5} annotations, on model \textbf{Accuracy} and \textbf{C}lass \textbf{Imbalance} for ATIS dataset.}
    \label{tab:noise-types}
\end{table}

We find that GPT-3.5 adds the most detrimental noise while having a similar class imbalance as other noise models.
We posit that GPT-3.5 provides very reasonable false annotations that are inherently hard to detect or ignore which leads to poor test performance.  
It is also interesting to note that a model trained on GPT-3.5-annotated training data performs much better than GPT-3.5 itself (84.0\% versus 75.6\%). 
This is due to the input data distribution learned by the model leading to better input representations compared to a generic pre-trained LLM. 
This motivates the fine-tuning of task-specific models for modular AI systems as well as the use of ALC to improve data quality which can lead to further improvements.

\begin{figure}
    \centering \includegraphics[width=\linewidth]{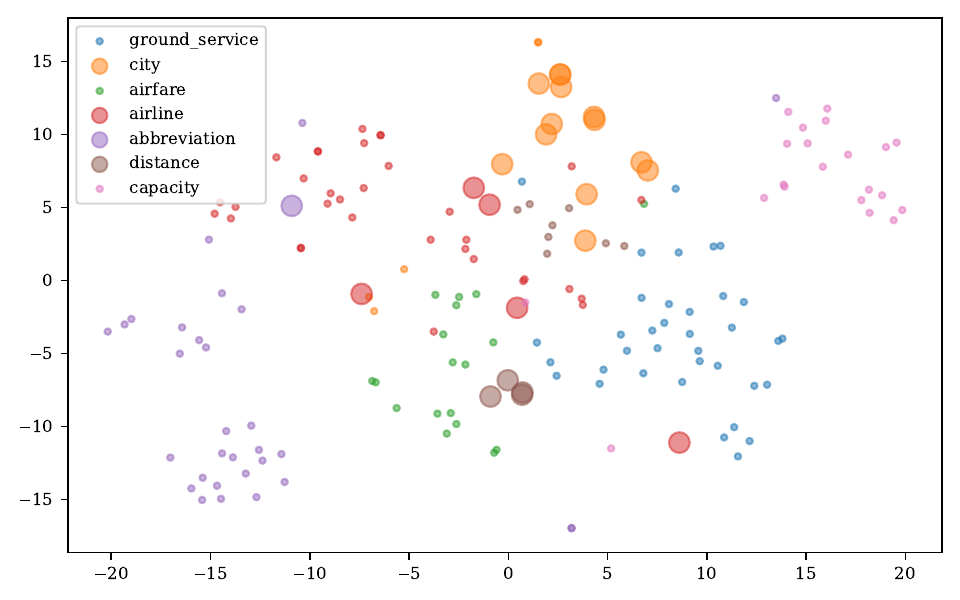}
    \caption{
    A 2D projection of ATIS text embeddings for a subset of 7 classes. 
    GPT-3.5 annotations are indicated by colors while large dots indicate errors.
    Most misannotated examples lie near cluster boundaries.
    }
    \label{fig:chatgpt-atis-projection}
\end{figure}

In Figure \ref{fig:chatgpt-atis-projection}, we show 2D t-SNE projections \cite{maaten_visualizing_2008} of Contriever text embeddings \cite{izacard_unsupervised_2022} from ATIS dataset.
We visualize examples labeled with 7 classes shown in the legend and larger dots indicate the examples misannotated by GPT-3.5. 
We observe that most misannotated examples lie near intersections or between class clusters.
This indicates that noise induced by GPT-3.5 is indeed complex and depends on the text, its true label, and other possible labels.
Previous work has used synthetic random \cite{kremer_robust_2018} or label-conditional noise \cite{rebbapragada_active_2012,li_improving_2022} which, as observed in Table \ref{tab:noise-types}, are not representative of noise present in GPT-3.5 annotations that we study. 

\subsection{Misannotation Prediction}

\begin{figure}[ht!]
    \centering
    \includegraphics[width=\linewidth]{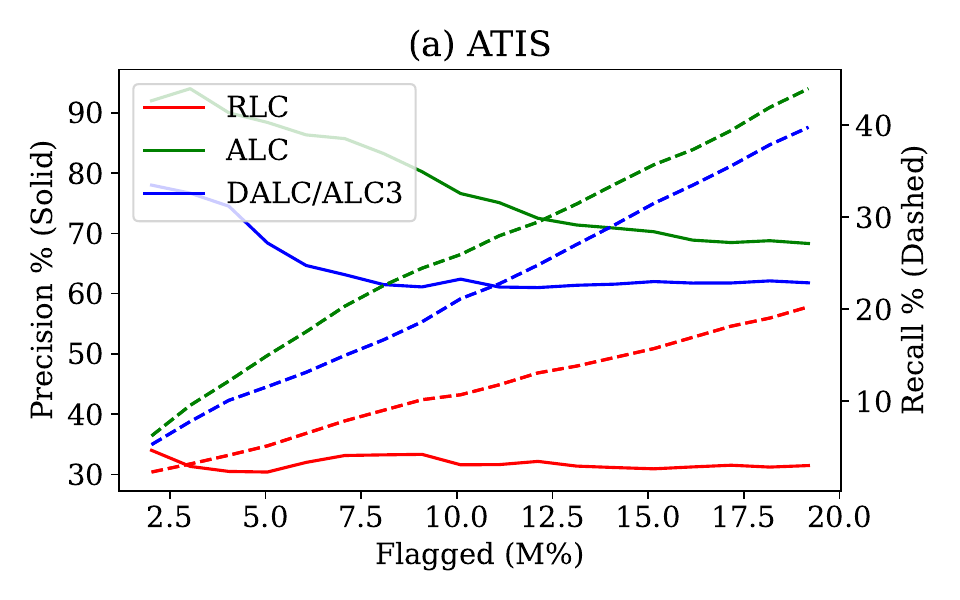}

    \includegraphics[width=\linewidth]{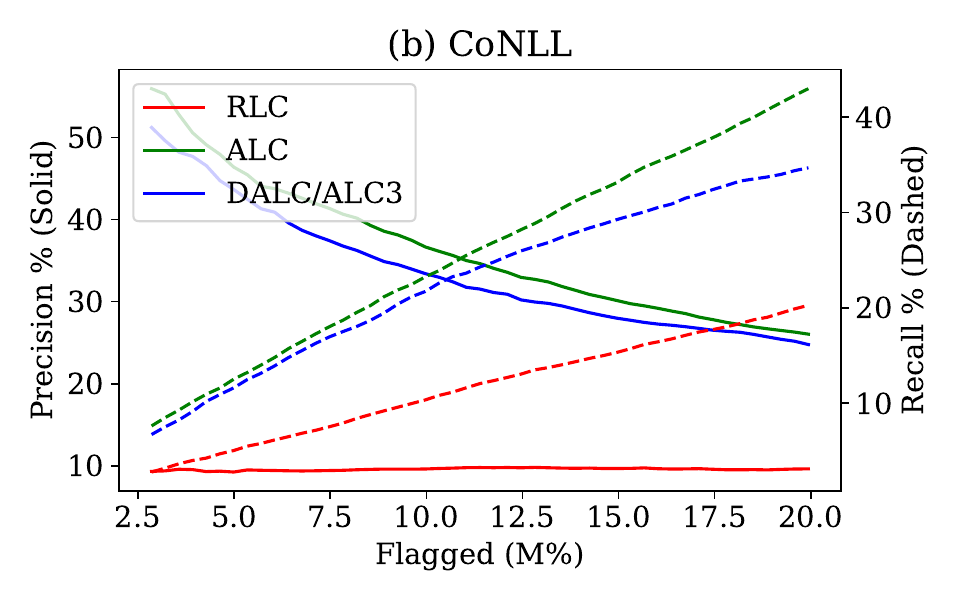}

    \includegraphics[width=\linewidth]{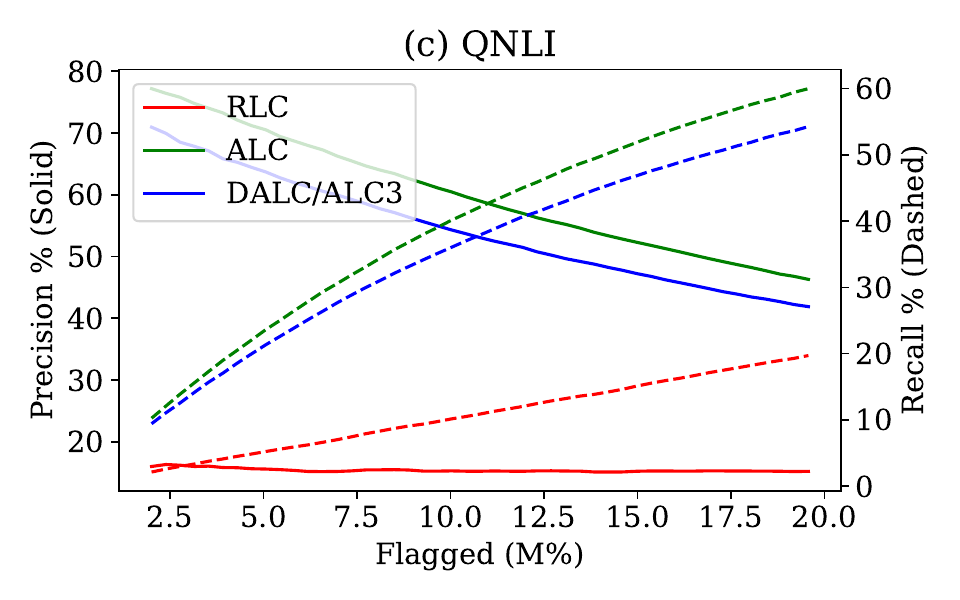}

    \caption{
    MP precision and recall for ATIS, CoNLL, and QNLI with change in $M$.
    ALC3 results are same as DALC, and both perform worse than ALC because data quality is improved with auto-correction before MP.
    }
    \label{fig:Methods}
\end{figure}

In Figure \ref{fig:Methods}, we compare \textbf{RLC}, \textbf{ALC}, and \textbf{DALC} for MP on all three datasets annotated using GPT-3.5. 
Note that filtering in \textbf{ALC3} doesn't impact misannotation prediction in the first iteration, leading to same results as \textbf{DALC}.
We use $\delta=0.75,0.98,0.90$ for ATIS, CoNLL and QLNI to keep the proportion of auto-corrected examples between 3\% to 6\%.
We note from Table \ref{tab:ChatGPTPerf} that 29.8\% of the ATIS and 15.1\% of the QNLI train dataset were misannotated by GPT-3.5. 
For CoNLL, 57.4\% of examples contain errors in token classes though only 9.7\% of tokens in the dataset are misclassified.

We observe that precision decreases for \textbf{ALC} and \textbf{DALC} as $M$ increases for all tasks
because the examples with the highest probability of being misannotated based on $m(.)$ values indeed have a higher rate of being misannotated. 
In other words, beginning from examples with the highest $m(.)$ values, a human annotator will see a decrease in MP quality over time as they verify examples with lower $m(.)$ values.
We also see an increase in recall as $M$ increases because more misannotated examples are flagged at higher values of $M$. 
We also note that \textbf{DALC} has consistently lower precision and recall than \textbf{ALC}.
The auto-corrected examples in \textbf{DALC} typically have high $m(.)$ as $p_\theta(y\neq y^*|x)$ is low but these examples cannot be flagged.  
\textbf{ALC} flags the examples with the highest $m(.)$ values leading to higher precision and recall.
For \textbf{RLC}, the precision is constant and low while recall increases perfectly linearly with $M$ because examples are flagged at random.

For the human annotator, examples flagged by \textbf{DALC/ALC3} are more likely to be correct as compared to those flagged by \textbf{ALC}.
A more balanced set of flagged examples for the correction task may also avoid the problem of over-reliance of human annotators on model predictions.
Further, increase in quality per iteration can be higher with \textbf{DALC/ALC3} because of two sources of true labels viz. auto-flipping and human corrections.
Given a method, we wish to maximize precision to correct a higher proportion of misannotated examples in each iteration. 
Therefore, we use low values of $M$ in each step in iterative data correction.

\subsection{Effect of Data Size on MP}

\begin{figure}[t]
    \centering
    \includegraphics[width=\linewidth]{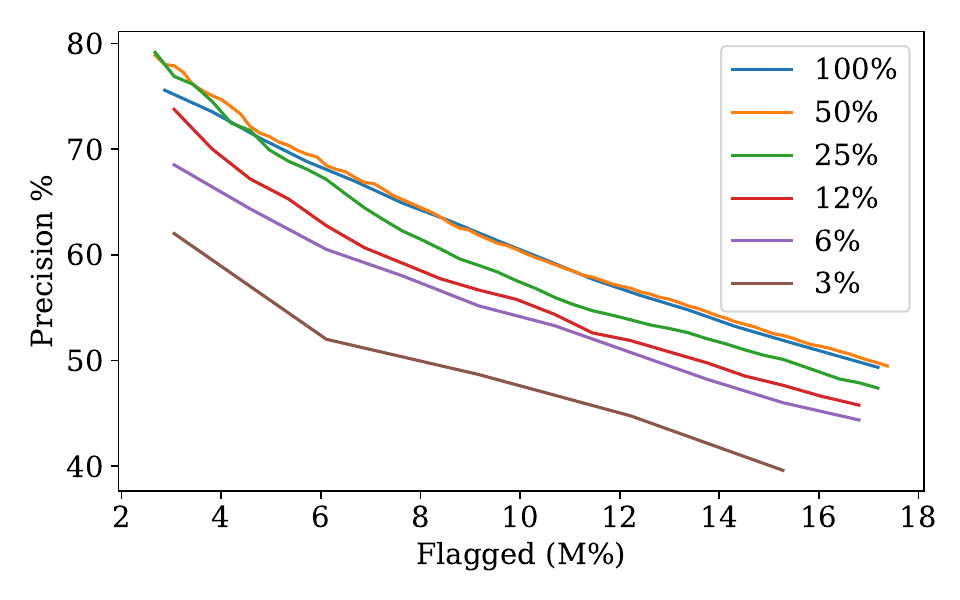}
    (a) MP precision versus flagged examples ($M$)
    \includegraphics[width=\linewidth]{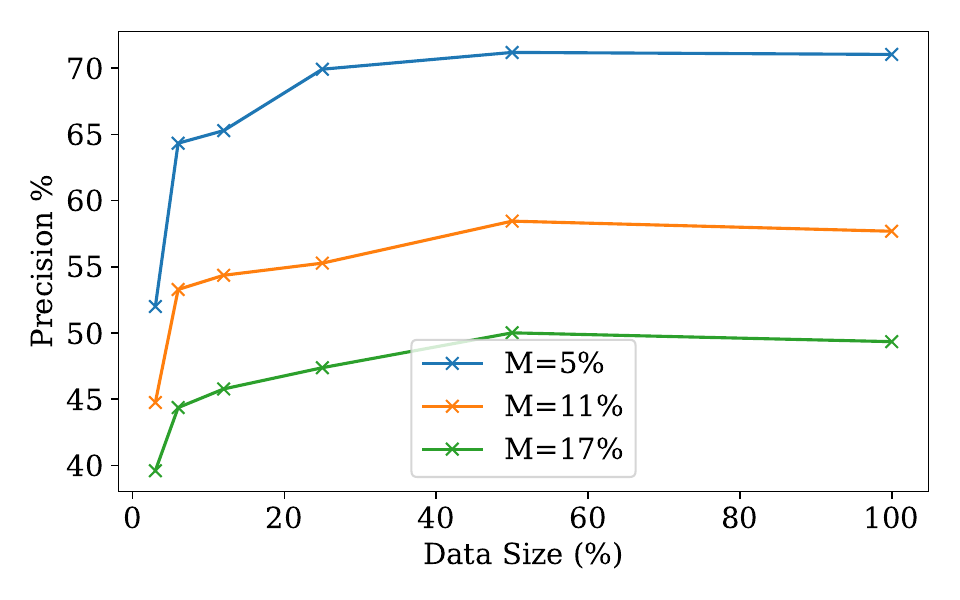}
    (b) MP precision versus data size (\%)
    \caption{Effect of data size on MP precision. MP precision reduces as more examples are flagged and reduces as data size is decreased, but we observe diminishing returns with increase in data size.}
    \label{fig:DataEffect}
\end{figure}

We wish to understand how the quantity of data collected by modules within an AI system will impact MP performance.
Towards this end, we vary the amount of data used to train the model for MP. 
We use 100\%, 50\%, 25\%, 12\%, 6\%, and 3\% of the QNLI dataset, the largest of the three with 105k examples (3\% $\approx$ 3.1k), to evaluate the impact of data size. 
The results are shown in Figure \ref{fig:DataEffect}.

{}

We observe a decrease in the MP precision as the data size decreases (Figure \ref{fig:DataEffect}(a)). 
The drop in precision from 6\% to 3\% is much bigger than drop from 100\% to 50\%
as each example plays a bigger role in learning for small data sizes.
In Figure \ref{fig:DataEffect}(b), we plot the precision as a function of data size with a fixed $M$. 
The precision increases as the data size increases, but with diminishing returns.
In practice, the observed precision can suggest the data size increase required to improve MP performance. 

\begin{figure}[t!]
    \centering
    \includegraphics[width=\linewidth]{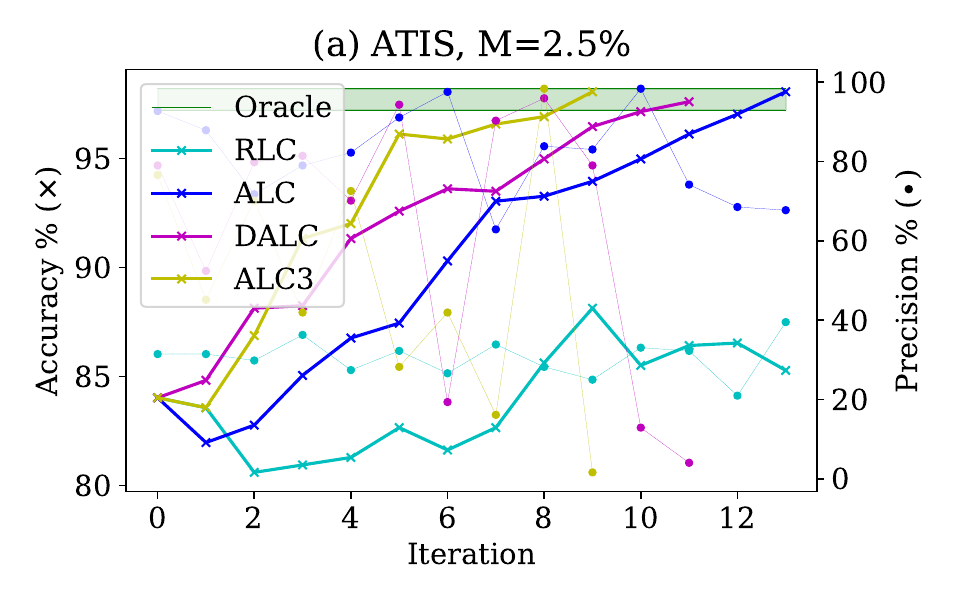}

   \includegraphics[width=\linewidth]{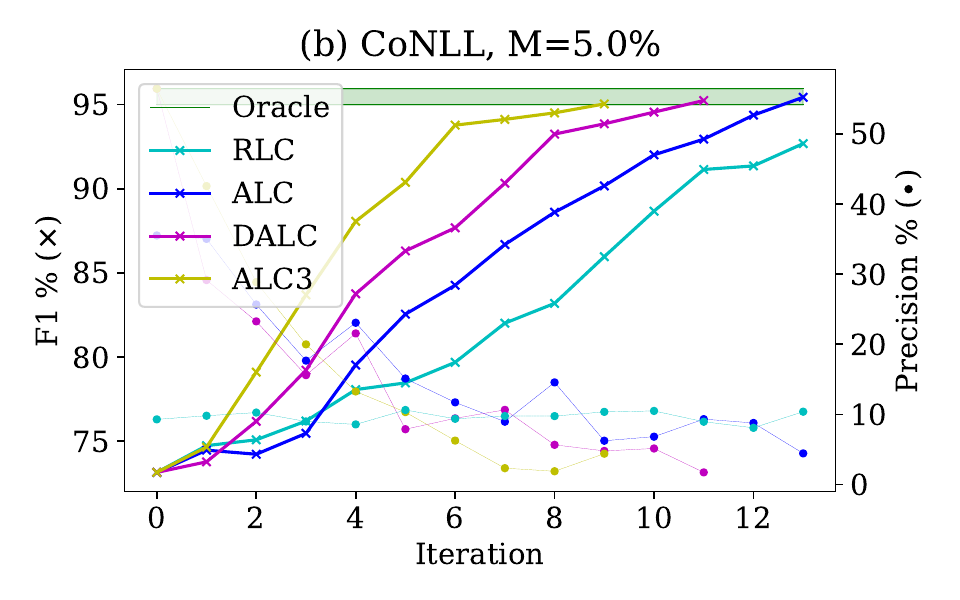}
   
   \includegraphics[width=\linewidth]{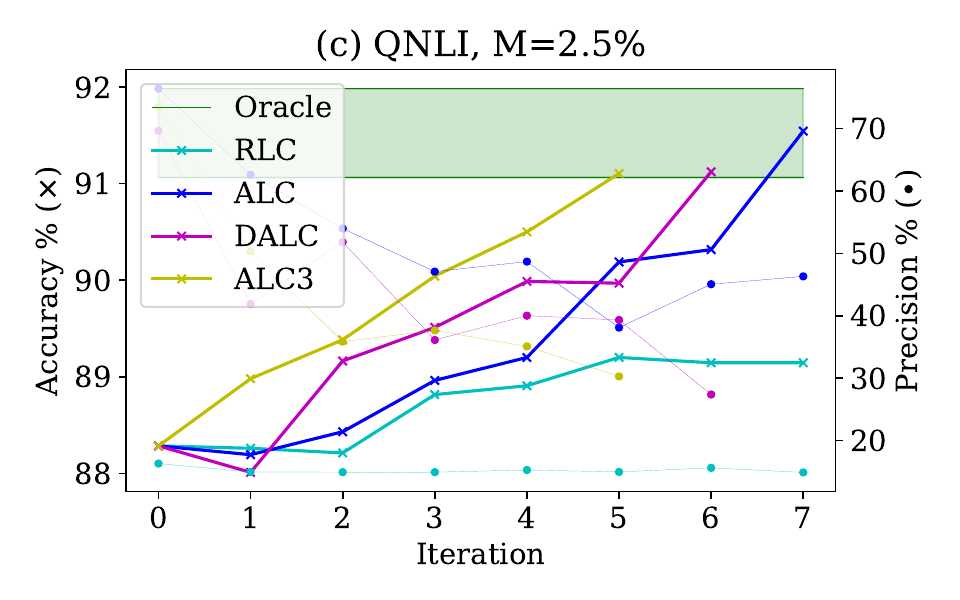}
   
   \caption{Model performance ({Accuracy/F1-score}) with iterations of simulated human verification using {RLC}, {ALC}, {DALC}, and ALC3 along with MP {Precision} and the {Oracle} performance.
    Accuracy/F1-score ({$\times$}) increases while
    MP precision ({$\bullet$}) decreases with each iteration. 
    We iterate until close-to-oracle performance is achieved i.e. Accuracy/F1 is within 1\% of the ground truth fine-tuned model (\textcolor{darkgreen}{\textbf{green bar}}).
    }
    \label{fig:AccVSVerified}
\end{figure}

\subsection{Iterative Data Correction}
\label{subsection:results-iterative}

The quality of training data improves after each iteration of label correction  
and the iterative process can be continued until we obtain a high-quality dataset.
We simulate correction, without actual human annotators, by using ground truth from the original dataset to update the annotations of the flagged examples.
We fix $M=0.025$ (2.5\%) for ATIS and QNLI tasks, but $M=0.050$ (5.0\%) for CoNLL because $> 50\%$ examples contain noisy tokens. 
For CoNLL task, we report the F1-score on token-level classification as a standard practice.
However, we predict misannotations on the sentence level since the human annotators verify complete sentences. 
Therefore, we calculate MP precision by treating every token as flagged if a sentence is predicted to be misannotated. 
While this artificially leads to low precision, it is more useful to observe compared to sentence-level precision which is consistently high because most examples for CoNLL contain misannotated tokens.

We train models with the updated datasets and plot performance i.e. \textbf{Accuracy} or \textbf{F1-score} ({$\times$} markers) after each iteration in Figure \ref{fig:AccVSVerified}.
We also plot the \textbf{MP precision} ({$\bullet$} markers) for each iteration. 
We say that a model has achieved close-to-oracle performance if the accuracy or F1-score is within a 1\% range of the model trained on the original dataset (\textbf{Oracle}).
The horizontal \textcolor{darkgreen}{\textbf{green bar}} in each plot shows this 1\% range.
For the ATIS task, ALC3, DALC and ALC achieve close-to-oracle performance in 9, 11 and 13 iterations respectively. 
With $M=2.5\%$, this means only $2.5\% \times 9 = 22.5\%$ of the dataset is human-annotated with ALC3 which is less than the $29.8\%$ of data misannotated by GPT-3.5. 
ALC3 requires about 18\% less annotations than DALC and 31\% less than ALC.
For CoNLL, we again observe that ALC3, DALC, and ALC require 9, 11, and 13 iterations to achieve close-to-oracle performance.
Since $57.4\%$ of CoNLL annotations by GPT-3.5 contain tokens with incorrect classes,
it is reasonable to human-annotate $5.0\% \times 9 = 45\% ~(< 57.4\%)$ of the data.
For QNLI, we need 5 iterations using ALC3 i.e. correcting $2.5\% \times 5 = 12.5\% ~ (< 15.1\%)$ of the dataset. 
For RLC, MP precision remains constant as examples are randomly selected for correction. 
It leads to slower improvements in  performance compared to other methods. 
With \textit{perfect MP}, we would require 29.8\%, 57.4\%, and 15.1\% examples for ATIS, CoNLL, and QNLI respectively to lead to the original dataset quality.
But ALC3 has close-to-oracle performance with 17-24\% fewer examples than perfect MP by using human feedback in addition to auto-correction and filtering.

We also observe a gradual decrease in MP precision for ALC, DALC, and ALC3 due to a decrease in the number of misannotated examples after each iteration.
As human annotators examine the flagged examples in practice, we can calculate the MP precision after each iteration and use it as a stopping criterion or as a signal to decrease $M$.
Additionally, as mentioned earlier, we hold out a human-annotated random test set to evaluate model performance. 
If the MP precision is no longer in a target range or the model performance saturates, one can stop performing iterations until more data becomes available from the deployed AI system.
Overall, we find that ALC3 can achieve close-to-oracle performance by correcting only a fraction of the dataset, 
17-24\% than the fraction of 
misannotated examples for all three GPT-3.5-annotated 
datasets, and $\approx 16\%$ less than DALC.

\section{Conclusion}

Modular AI systems can leverage LLMs as zero/few-shot learners for rapid deployment without any training or fine-tuning but there is a large performance gap when compared to fine-tuned models. 
The proposed pipeline utilizes our method ALC3 to improve modular AI systems by leveraging LLM-annotated data and improving its quality to train a replacement model for each module.   
Compared to random or label-conditioned synthetic noise models used in previous work, we found the noise characteristics of GPT-3.5-annotated text data to be far more complex. 
With ALC3, we achieve the same performance as ground truth data using human feedback on a fraction of the dataset.  
We found this fraction to be 17-24\% less than the fraction of misannotated data for three unique tasks spanning different levels of complexity and data sizes, and $\approx16\%$ better than previous methods.

\section{Limitations and Future Work}

Most publicly available datasets have small text lengths while real applications may involve much longer (conversational) texts. Deploying an AI system and collecting real datasets is outside the scope of this paper. 
In future work, we will apply ALC3 to datasets created using data traces obtained from a deployed Jill Watson \cite{kakar_jill_2024} to evaluate its performance after replacing LLM calls with smaller fine-tuned models.

Most commercial conversational AI agents allow users to provide feedback through a thumbs-up/down button and 
this feedback may be accommodated as a signal for misannotation prediction. 
We leave this as an open problem for future work.

We used only GPT-3.5 as LLM for annotations. While our results should generalize to different LLMs since this only affects the initial parameters of our experiments (i.e. the noise level of the labelled datasets), more insights could be gained by testing with other LLMs.

\bibliography{references}

\newpage

\appendix

\section{Extension to Generative Models} \label{appendix:extension-generative}

To generalize our approach to generative tasks, we can use the conditional probability of generated text given a text input. 
$$ P(y|x) = P(y_1,\ldots,y_n|x) = P(y_{1:n}|x) $$

For decoder models, sampling is performed auto-regressively by assuming $P(y_1,\ldots,y_n|x) = P(y_1|x) \cdot P(y_2|y_{1},x) \ldots P(y_{n}|y_{1:n-1},x)$.
After model training with a dataset of input and output texts, the probability of misannotation can be calculated as usual.
$$m(x, y_{1:n}) = 1 - P(y_{1:n}|x)^{\frac{1}{n}}$$

Since generative tasks are far more complex than discriminative tasks, 
one can expect that the amount of data required to train a model for misannotation prediction for generative tasks will be much larger than data size required for discriminative tasks.
Further, obtaining human feedback or creating a reasonable proxy for human feedback for generative tasks is another difficult challenge.
In future work, we will examine the data size required to effectively perform annotation prediction for different generative tasks.

In previous work in human-in-the-loop learning for generative tasks, Reinforcement Learning with Human Feedback (RLHF) used for ChatGPT \cite{openai_introducing_2022} provides an answer to a part of the problem 
but doesn't address how to select useful examples where human-feedback will be most helpful in terms of improving model performance.

\section{Training Details}\label{appendix:training-details}

The training details for models for each task are provided in Table \ref{tab:appendix_training_details}. 

\begin{table*}
    \centering
  \begin{tabular}{lccc}
    \toprule
    \multicolumn{1}{l}{\textbf{}} & \textbf{ATIS} & \textbf{CoNLL} & \textbf{QNLI}\\
    \midrule
    Model Name & distilbert-base-uncased & roberta-base & albert-large-v2 \\
    Trainable Parameters & 67M & 124M & 17M \\
    Training examples  & 4,952 & 14,041 & 104,743 \\
    Epochs & 9 & 6 & 3 \\
    Batch Size & 16 & 16 & 16 \\
    Optimizer & AdamW & AdamW & AdamW \\
    Learning Rate & 1e-4 (constant) & 2e-5 (constant) &  1e-5 (constant) \\
    Warmup Steps & 155 & 439 & 0 \\
    \bottomrule
  \end{tabular}
    \caption{Training details for ATIS, CoNLL and QNLI tasks.}
    \label{tab:appendix_training_details}
\end{table*}

\section{GPT-3.5 Prompts for Evaluation Tasks}\label{appendix:prompts}

\subsection{ATIS}

ATIS (Airline Travel Information System) Intent  Classification dataset \cite{hemphill_atis_1990} is labeled with 17 unique intents with one or more intents assigned to user queries. 
There are only 40 examples with multiple labels. We removed them from the dataset leading to a single-label classification task.
We provided GPT-3.5 with the system message below to instruct it for classification.

\vspace{5px}
\noindent\fbox{
\parbox{0.95\linewidth}{
`role': `system'

`content': `Your task is to classify the query into possible classes.'
}}
\vspace{5px}

Along with the system message, we provide the user message to describe the problem, the input text, and the output format.

\vspace{5px}
\noindent\fbox{
\parbox{0.95\linewidth}{
`role': `user'

`content': `Classes:

~~[Class 1]: [class description],

~~...

~~[Class 17]: [class description], 

~~Query:

~~[input text]

~~Answer exactly one of [class names separated by commas].'
}}
\vspace{5px}

A few examples of classes and their descriptions are as follows:

\begin{enumerate}
    \item \textbf{`aircraft'}: `This intent is associated with queries that seek information specifically about an aircraft, such as its features, type, or specifications.'
    \item \textbf{`flight'}: `This intent covers general queries related to flights, including information about schedules, availability, or any other flight-related inquiries.'
    \item \textbf{`flight time'}: `Queries with this intent typically seek information about the duration or specific timing of a flight.'
    \item \textbf{`cheapest'}: `This intent indicates queries that focus on finding the most economical or low-cost options, such as the cheapest flights or fares.'
    \item \textbf{`ground fare'}: `Queries with this intent are related to ground fares or transportation costs, such as taxi fares or shuttle services.'
\end{enumerate}

To retrieve the intent from the output, we iterate over the intents and use the first found intent in the output text.

\subsection{CoNLL}

To extract named entities from examples in CoNLL 2003 dataset \cite{tjong_introduction_2003}, we provided GPT-3.5 with one example of entity recognition output to describe the output pattern. 
First, we used the user message below to provide the types of entities that need to be extracted.

\vspace{5px}
\noindent\fbox{
\parbox{0.95\linewidth}{
`role': `system'

`content': `Your task is to extract named entities of type Person, Location, Organization and Miscellaneous.'
}}
\vspace{5px}

Next, we use the user message below to provide an example and the input text from which entities need to be extracted. We use the same NER example below for every input.

\vspace{5px}
\noindent\fbox{
\parbox{0.95\linewidth}{
`role': `user'

`content': `Example input:

~~``Mercury was born Farrokh Bulsara in Stone Town in the British protectorate of Zanzibar (now part of Tanzania) on 5 September 1946 ."

~~Example output:

~~``[Mercury](Person) was born [Farrokh Bulsara](Person) in [Stone Town](Location) in the [British](Miscellaneous) protectorate of [Zanzibar](Location) (now part of [Tanzania](Location)) on 5 September 1946 ."

~~Input:

~~[input text]'
}}
\vspace{5px}

To parse the output, we use regex to extract entity types and align words in the input and output to get classification for each token.

\subsection{QNLI}

QNLI (Question-Answering Natural Language Inference) dataset \cite{wang_glue_2018} has pairs of questions and passages based on the Stanford Question Answering Dataset and the label $0$ if the passage answers the question and $1$ otherwise. 
We use the system message below to prime GPT-3.5 with the task. 

\vspace{5px}
\noindent\fbox{
\parbox{0.95\linewidth}{
`role': `system'

`content': `Your task is to solve a textual entailment task.'
}}
\vspace{5px}

Next, we use the following user message to give the input question and passage and prompt GPT-3.5 to say `YES' if the passage can answer the question and `NO' otherwise.

\vspace{5px}
\noindent\fbox{
\parbox{0.95\linewidth}{
`role': `user'

`content': `Question:[input question] 

~~Passage: [input passage]

~~Can the given passage be used to answer the question?

~~Answer exactly YES or NO (capitalized).'
}}
\vspace{5px}

Finally, we label the example with $0$ if `YES' is present in the output and $1$ otherwise. 

\section{Misannotation Prediction for CoNLL}

For CoNLL, each sentence is provided with a list of tokens and token-level classification.
But we perform misannotation prediction (MP) on the sentence level since the human annotators evaluate the complete sentence and assign named entity types based on the context. 
Therefore, we use combine the probabilities of token-level misannotations into a sentence-level misannotation probability. 
Let $j$th token of sentence $x_i$ be $w_j$ with named entity label $y_i^{w_j}$. Let there be $L_i$ tokens in the sentence $x_i$.  

An example is correctly annotated if all the tokens in the sentence are correctly labeled. Therefore, the probability of correct sentence-level annotation is the product of token-level probabilites.
$$p_{\theta}(y_i|x_i) =  \prod_{1 \leq j \leq L_i} p_{\theta}(y_i^{w_j}|x_i) $$ 

But leads to a lower probability for longer sentences. To normalize for length, we use the following equation:

$$p_{\theta}(y_i|x_i) = \Big( \prod_{1 \leq j \leq L_i} p_{\theta}(y_i^{w_j}|x_i) \Big) ^ {1/L_i}$$ 

which is the geometric mean of token-level probabilities.

\section{Noise Models}\label{appendix:noise-models}

For \textbf{random noise}, we randomly select examples for the noising process. 
For each selected example, we change the label to any other label with equal probability.
In other words, we assume that each example is equally likely to be mislabeled and each label is equally likely to be replaced by another label.
\cite{kremer_robust_2018} uses this process to add noise to clean datasets for experimentation.

For \textbf{label-conditional noise}, we first train a classification head on top of the BERT model (`bert-base-uncased' on HuggingFace) and use its probability outputs on training examples to calculate label transition probabilities.
For each label, we calculate the transition probabilities by averaging the output distribution of examples with that label.
After this, we randomly select examples for noising and change their labels based on the transition probabilities.
This process assumes that each example is equally likely to be noisy but noisy label is conditionally dependent on the true label.
\citet{li_improving_2022} use the same process for introducing noise into clean datasets in their experiments. 

For \textbf{input-conditional noise}, we use the same BERT + fine-tuned classification head model to calculate the output probabilities of each example.
We iterate over the dataset to sample new labels for each example based on its output probability distribution.
We iterate over random permutations of the dataset multiple times but only sample new labels for unchanged examples.
This is repeated until the required number of noisy examples is achieved.
In this process, an example is likely to be mislabeled with probability proportional to $1-p_{\theta}(y_i|x_i)$.
Further, the new label is sampled with probability proportional to $p_\theta(y \neq y_i|x_i)$.

\end{document}